\documentclass[letterpaper, 10 pt, conference]{ieeeconf}  

\IEEEoverridecommandlockouts                              

\usepackage{graphics} 
\usepackage{epsfig} 
\usepackage{mathptmx} 
\usepackage{times} 
\usepackage{amsmath} 
\usepackage{amssymb}  
\usepackage{gensymb}
\usepackage{hyperref} 
\usepackage{multirow}
\usepackage{hhline}
\usepackage{algorithm}  
\usepackage{algpseudocode}  
\usepackage{cite}
\usepackage{subfigure}       
\usepackage{graphicx}
\usepackage{color}
\usepackage{xcolor}

\title{\LARGE \bf
Descriptor Distillation for Efficient Multi-Robot SLAM
}

\author{Xiyue Guo$^{1}$, Junjie Hu$^{2}$, Hujun Bao$^{1}$ and Guofeng Zhang$^{1\dagger}$
 \\$^{1}$State Key Lab of CAD\&CG, Zhejiang University \quad  $^{2}$Chinese University of Hong Kong, Shenzhen
\thanks{$^{\dagger}$Corresponding author.
        Email: {\tt\small zhangguofeng@zju.edu.cn}
        }%
\thanks{This work was partially supported by NSF of China~(No. 61932003).}
}
\begin{document}

\maketitle
\thispagestyle{empty}
\pagestyle{empty}

\begin{abstract}
Performing accurate localization while maintaining the low-level communication bandwidth is an essential challenge of multi-robot simultaneous localization and mapping (MR-SLAM). In this paper, we tackle this problem by generating a compact yet discriminative feature descriptor with minimum inference time. We propose descriptor distillation that formulates the descriptor generation into a learning problem under the teacher-student framework. To achieve real-time descriptor generation, we design a compact student network and learn it by transferring the knowledge from a pre-trained large teacher model. To reduce the descriptor dimensions from the teacher to the student, we propose a novel loss function that enables the knowledge transfer between two different dimensional descriptors. The experimental results demonstrate that our model is 30\% lighter than the state-of-the-art model and produces better descriptors in patch matching. Moreover, we build a MR-SLAM system based on the proposed method and show that our descriptor distillation can achieve higher localization performance for MR-SLAM with lower bandwidth.

\end{abstract}

\section{INTRODUCTION}

Multi-robot simultaneous localization and mapping (MR-SLAM) aims at perceiving environments by utilizing a set of cooperative robots \cite{CCM,covins,DoorSLAM,XiyueGuo2021}. It is an extended approach to single-robot SLAM (SR-SLAM) and has attracted significant attention due to its clear advantage in exploring large-scale environments. However, a multi-robot system brings one additional restriction: the limitation of the communication bandwidth. In a MR-SLAM system, to integrate all the trajectories from the whole team, every robot in the team should share its keyframe data (including keyframe pose and observed feature points) to process the inter-robot loop-closure and global localization. 
This type of data exchange occupies a high communication capacity, which is very likely to degrade real-time performance. Some engineering solutions, such as reducing the communication frequency, and declining the keypoints number, are able to reduce the bandwidth. On the other hand, these approaches cause another problem: the localization results in lower accuracy. 

In order to maintain high localization performance while reducing the bandwidth, producing more compact features is a reasonable and universal solution. To this end, some previous approaches~\cite{narrow1}~\cite{narrow2} have attempted to reduce the dimension of the handcraft descriptors. However, this kind of methods leads to poor matching performance.

\begin{figure}[t]
\centering
\includegraphics[width=2.5in]{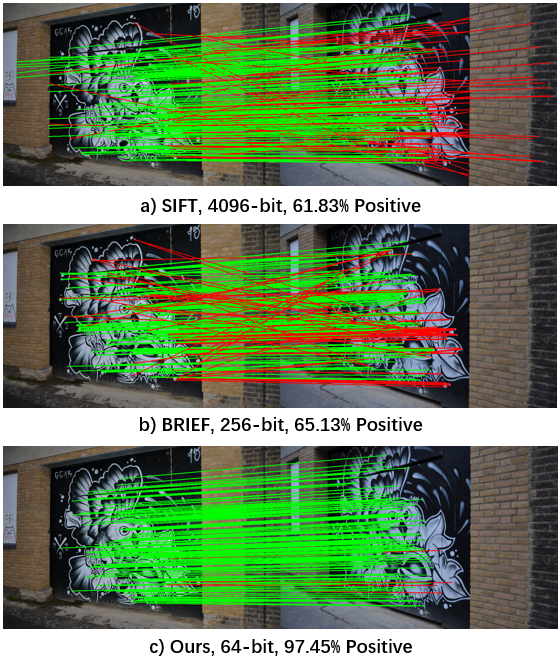}
\vspace{-2mm}
\centering
\caption{A visualization comparison of generated descriptors between previous methods and our method. As seen, our method extracts significantly compact descriptors and achieves more accurate matching performance.}
\label{fig:matching}
\vspace{-5mm}
\end{figure}

Recently, convolutional neural network (CNN) approaches~\cite{L2-net,hardnet,SOSnet,HyNet} have shown their superior performance against handcraft descriptors. 
However, these methods are specially designed and evaluated only for SR-SLAM and have no guarantee to be applied to MR-SLAM.
Typically, there are two difficulties in applying CNN-based descriptors to MR-SLAM. First, most state-of-the-art approaches are designed to produce high-dimensional (more than 128 dimensions) descriptors. 
Second, it is hard for most current CNN models to perform real-time descriptor generation in the mobile platform since they tend to use large and complex networks. 
The high model complexity and descriptor dimensions are two major obstacles that hinder the deployment of learning-based methods to MR-SLAM.

\begin{figure*}[t!]
\centering
\includegraphics[width=5.5in]{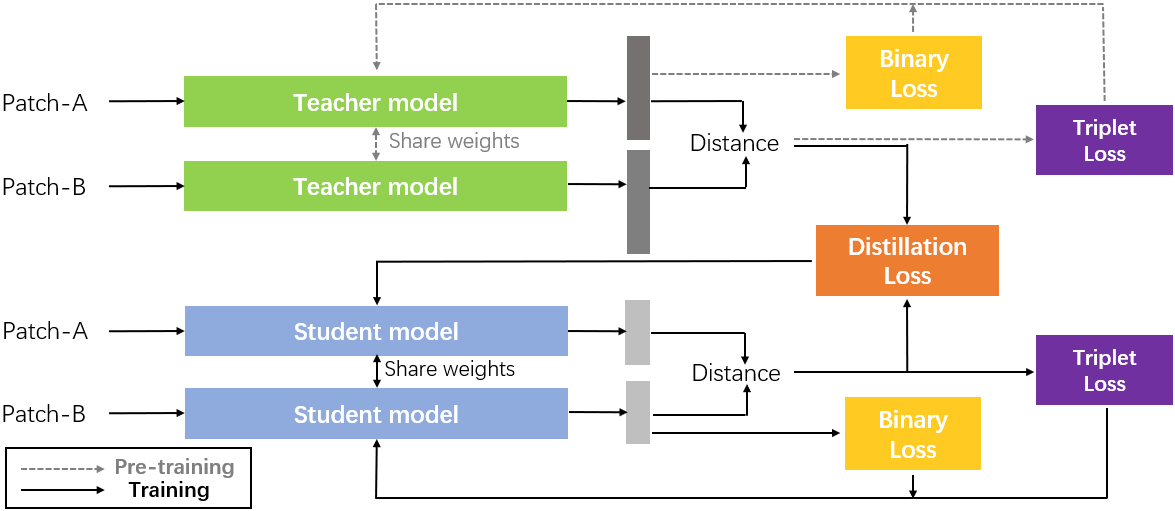}
\vspace{-0mm}
\centering
\caption{The diagram of the proposed teacher-student framework for descriptor distillation. 
The teacher model is firstly trained with common losses, and its parameters are fixed during student learning. Then, the high-dimensional descriptors predicted by the teacher are utilized to guide the student to produce low-dimensional descriptors with the proposed distillation loss. Therefore, the loss function of the student model consists of several common losses as well as the distillation loss.
}
\label{fig:framework}
\vspace{-3mm}
\end{figure*}

In this paper, we aim to design a compact network that can produce low-dimensional descriptors with a small number of parameters. Most importantly, the generated descriptors have to outperform or at least be comparable to those predicted by using larger models. 
Towards this goal, we propose a knowledge distillation (KD) based framework for descriptor learning. Our framework is a classic teacher-student combination, where the student is the compact model which we aim to learn, and the teacher is a pre-trained larger network whose predictions are used for supervising the student learning.
However, KD requires the same dimensional predictions from the teacher and the student to calculate a loss function \cite{Hinton2015DistillingTK,hu2021boosting,wang2021knowledge}.
In our problem, predicted descriptors between the teacher and the student have different dimensions, making KD incompatible with descriptor distillation.

To address this problem, we propose a novel distillation loss function that can distill feature descriptors between the two models even if they yield different dimensional outputs. To do this, we 
compute the similarity between the negative pairs (unmatched pair) distances across the teacher and student models.
Based on this distillation approach, our network is able to alleviate the performance drop while reducing both model parameters and descriptor dimension. Fig.~\ref{fig:matching} shows an example comparison between previous traditional methods and our method. We fairly test our models on public patch matching datasets. Moreover, based on our descriptor model, we develop a MR-SLAM system to quantify the performance of our descriptors on the real-world SLAM task.  




To summarize, we present the following contributions:
\begin{itemize}
\item
We design a teacher-student model to generate the compact binary descriptor. Our model outperforms state-of-the-art methods in both matching performance and computational cost.

\item 
We propose a novel distance-based distillation loss, which allows knowledge transfer  between models with different output dimensions.

\item
We develop a MR-SLAM system based on the new descriptor model as the evaluation platform. The results show that our model demonstrates good performance for MR-SLAM task in public EuRoC dataset~\cite{euroc}. 
\end{itemize}


\section{RELATED WORK}
\subsection{Bandwidth narrowing approaches}
Recently, most MR-SLAM systems share all keyframe data between robots or servers. This data exchange mode will easily exceed the bandwidth limitation during real-world tasks. Moreover, simply reducing the containing information or sending frequency of the keyframes will bring poor localization performance. To deal with the difficulty, Some works only share the compact global descriptors for loop-closure~\cite{DoorSLAM}~\cite{talk}~\cite{near} in normal times. They exchange the whole keyframe data only when the loop is detected. However, this type of approaches is not suitable for all multi-robot architectures. Some other works use high-level information such as semantics to generate object-level landmarks for inter-robot localization~\cite{object1}~\cite{object2}. Since the object-level landmarks are much less than local feature points, these approaches can significantly reduce communication bandwidth. However, generating such high-level landmarks requires specific environments that contain rich semantic information. Some approaches try to deal with local features to maintain the generalization while reducing the bandwidth. \cite{narrow1} uses visual BoW indexes to replace the normal feature descriptors. In our paper, instead of compressing the existing descriptors, we explore how to generate the compact descriptors directly.

\begin{figure*}[t!]
\centering
\includegraphics[width=5.5in]{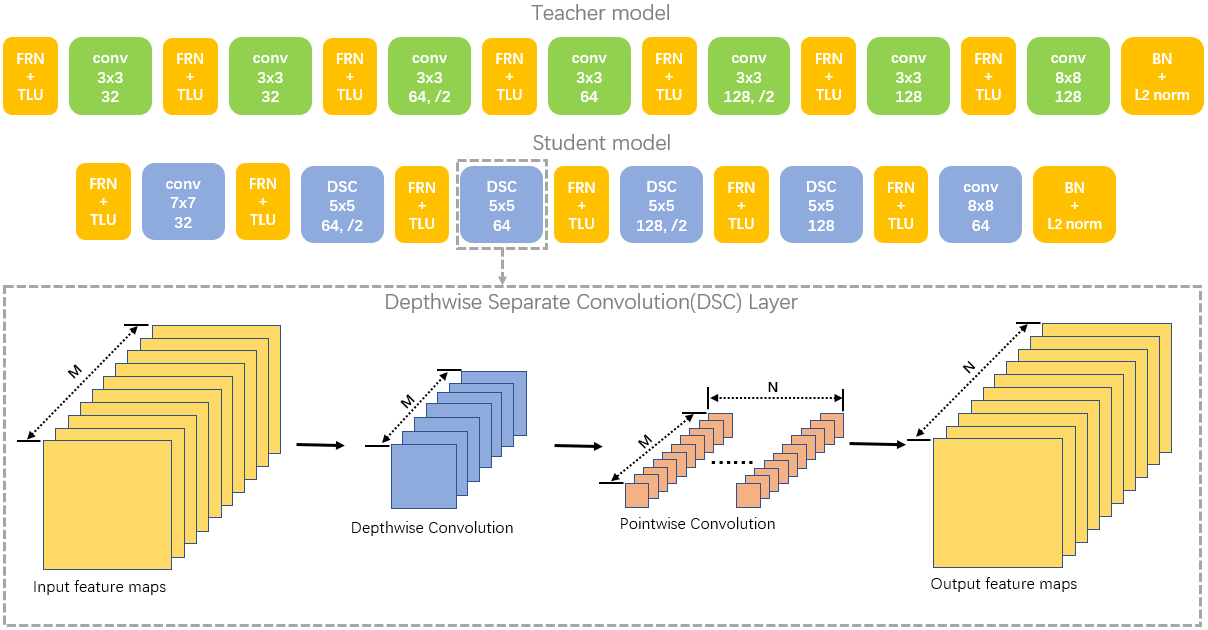}
\vspace{-3mm}
\centering
\caption{The detailed network architecture of the teacher model and the student model. The teacher model contains 7 convolution layers, while our student model has 4 depthwise separate convolution layers and 2 convolution layers (the first and last layers)}
\label{fig:model}
\vspace{-4mm}
\end{figure*}

\subsection{Traditional descriptors}
The goal of descriptors works is embedding the patch information into a vector space so that the corresponding patches have similar descriptors and vice versa. Early handcraft descriptors are mainly extracted from low-level information, such as gradient or intensity. SIFT~\cite{SIFT} describes the patches by gradient histograms. It has good matching performance but contains a large data size. BRIEF~\cite{BRIEF} has a more compact size due to its structure of binary vectors. However, its performance is decreased compared with SIFT. 
In order to obtain higher quality and more compact descriptors, several learning-based methods have been proposed. PCA-SIFT~\cite{PCASIFT} exploits principal components analysis (PCA) to generate the dimensional reduction version of SIFT. LDAhash~\cite{ldahash} utilizes the linear discriminant analysis (LDA) method to binarize the real-valued descriptor. BinBoost~\cite{binboost} learns 64-bits binary descriptors through the boost hash function. However, these learning methods are still based on low-level information; hence huge patch information is lost during the generation process. 

\subsection{CNN-based descriptors}
CNN-based approaches are end-to-end methods that generate the descriptors directly from the patches. In most works, Siamese architectures and triplet loss are utilized to learn descriptors~\cite{matchnet}. During the training process, distances between matched pairs are shortened, and unmatched pairs are lengthened. L2Net~\cite{L2-net} first generates the descriptors that can be matched by Euclidean space. Binary descriptors can be obtained by taking the sign of the real-valued vectors. To further improve the matching quality, HardNet~\cite{hardnet} only takes several closet unmatched pairs for training. SOSNet~\cite{SOSnet} adds a second-order-similarity as a regularization term into the training. Based on previous methods, HyNet~\cite{HyNet} proposed a hybrid similarity measurement method to measure the distance between descriptors during the training. 
On the other hand, some works also provide unsupervised models for descriptor learning~\cite{unsuper1, unsuper2, unsuper3}. However, the matching performances of these works are much less than the supervised methods. 
Although the recent CNN methods achieve excellent performance, the models and descriptors sizes of them are still too large for multi-robot systems.
Hence, In our paper, we follow the state-of-the-art supervised CNN methods and provide a more lightweight model to produce more compact descriptors for MR-SLAM.

\section{Methodology}
In this section, we present the proposed method in detail. Specifically, we first introduce preliminary knowledge of learning-based descriptor generation. Then, we provide our teacher and student architecture. Based on that, we present the proposed distillation loss, which is the key to the success of training the student. Finally, we give a detailed learning objective to train the student. The framework of our descriptor distillation is shown in Fig.~\ref{fig:framework}. 


\subsection{Preliminary}
\subsubsection{Siamese architecture}
Currently, mainstream CNN methods learn descriptors through the Siamese architectures. It contains twin models that share the same weight during the training. 
In the Siamese framework, two batches are prepared on every epoch of training. Each patch in a batch is negative to the other. On the other hand, the two batches in one epoch are one-to-one correspondence. There are two major losses used for training as follows.
\subsubsection{Triplet loss}
Triplet loss $L_{T}$ has been widely used in descriptor generating. It enforces the matched pairs are as close as possible while unmatched pairs are as far as possible. Most recent methods follow the HardNet strategy~\cite{hardnet}, which only extracts the negative pair that contains the smallest distance for training:
\begin{equation}
L_{T}=\sum_{i}^{N} \max \left(0,t+dis\left(R_{i}, R_{p}\right)-dis\left(R_{i}, R_{n}^{*}\right)\right),
\end{equation}
 In the equation, $R_{i}$ is the current real-valued descriptor, $R_{p}$ is the corresponding positive descriptor, $R_{n}^{*}$ is the closest negative descriptor, $N$ is the batch size. $dis$ calculates the $L_{2}$ distance between two descriptors. 
\subsubsection{Binarization loss}
Binarization loss term $L_{B}$ contributes to reducing the information loss when binarizing the real-valued descriptors. It aims to minimize the difference between the real-valued descriptor and the corresponding binary descriptor, which is represented as:
\begin{equation}
L_{B}= \sum_{i}^{N} \sum_{k}^{D} \frac{1}{D}\sqrt{\left(\left(R_i(k)\right)-B_i(k)\right)^{2}},
\end{equation}
where $B(k)$ is the $k$-th value of the binary descriptor $B$, and the $R(k)$ is the {\color{black}$k$-th
} value of the real-valued descriptor $R$. In addition, $D$ is the dimension of the descriptors.

\subsection{The Teacher-Student Distillation Framework}
Following the typical KD strategy ~\cite{Hinton2015DistillingTK,hu2022data,hu2022progressive,wang2021knowledge}, we exploit two siamese networks to form our teacher-student framework. The architecture of our network is given by Fig.~\ref{fig:model}. Our teacher architecture is adopted from HyNet~\cite{HyNet}. It consists of 7 convolutional layers and outputs 128-dimensional descriptors. Except for the final one, all convolutional layers are followed by Filter Response Normalisation (FRN) and Thresholded Linear Unit (TLU). Batch Normalization (BN) and L2-Normalization are placed after the final convolutional layers. 

For our student model, the output dimension is 64. We employ a more shallow network (6 convolutional layers) to reduce the time consumption. Furthermore, following the strategy of MobileNet~\cite{Mobilenets}, we replace the standard convolutional layers with depthwise separate convolution layers (DSC), except for the first and last one. The basic DSC layer consists of two parts. The first part is a depthwise convolution, it applies one convolution filter for each input channel. The second part is a 1$\times$1 convolution called pointwise convolution. 

Based on the architectures, each model is able to embed the image patches into the real-valued descriptors. To obtain the binary descriptors, we use the following equations to binarize the real-valued descriptors:
\begin{equation}
B(k)= \begin{cases}-1 & \text { if } R(k) \leq 0 \\ 1  & \text { if } R(k)>0\end{cases},
\end{equation}
where $B(k)$ is the $k-th$ value of the binary descriptor, and $R(k)$ is the $k-th$ value of the real-valued descriptor that is output from the model.

\subsection{Distillation Loss Function}
Due to the difficulty of transferring knowledge between two models with different dimensional outputs, we design a distance-based distillation loss function. It consists of a real-valued term and a binary-valued term, which enforce the descriptor batch from the student model to have a similar distribution to the teacher model on both real-valued and binary spaces, respectively. To be specific, it minimizes the difference in distance estimation between the same pairs by the teacher and student models. In addition, since the distances of the positive pairs are already extremely small, the differences between them are not worthy of reduction. Hence, we only minimize the negative pairs during the training. 

As mentioned above, the real-valued term minimizes two models' difference in the real-valued space, which can be formulated as:
\begin{equation}
L_{Real} = \sum_{i}^{N} \sum_{n}^{N_{n}}\frac{1}{N_{n}}\sqrt{(\lambda_{r}dis(R_{i}^{t}, R_{n}^{t}) -  dis(R_{i}^{s}, R_{n}^{s}))^{2}}
\end{equation}
where $N_{n}$ is the total number of negative descriptors relative to the current descriptor. $R_{i}^{t}$ and $R_{i}^{s}$ are the current real-valued descriptors that produced by the teacher and student, respectively, while the $R_{n}^{t}$ and $R_{n}^{s}$ are the corresponding negative descriptors. Moreover, due to the different dimensions between the teacher and student, we add a coefficient $\lambda _{r}$ {\color{black} that is set to 0.95 
} to adjust the scale of teacher-side distance.

Binary-valued term transfers the knowledge in binary space. It can be formulated as follows:
\begin{equation}
L_{BIN} = \sum_{i}^{N} \sum_{n}^{N_{n}}\frac{1}{N_{n}}\sqrt{(\lambda_b(B_{i}^{t} \cdot B_{n}^{t}) - ((B_{i}^{s} \cdot B_{n}^{s}) )^{2}},
\end{equation}
where $\lambda_b$ is a scaling coefficient that is usually set to the ratio between the output dimensions of the two models: ${D_s}/{D_t}$.
Note that in the training stage, we obtain the binary descriptor $B_{i}$ with the following representation:
\begin{equation}
B_{i} = \frac{R_{i}}{abs(R_{i})+\epsilon},
\end{equation}
where $\epsilon$ is a coefficient that prevents division by zero and is set to 1e-5 in our experiments. 

Finally, we combine the two loss terms as our distillation loss function, which is represented as:
\begin{equation}
L_{distillation} = L_{Real} + \gamma L_{Bin},
\end{equation}
where the $\gamma$ is weighting coefficient of the binary-valued term.

\subsection{The Student Training}
Our training framework consists of two stages. In the first stage{\color{black}, we train the teacher model using the ADAM optimizer with a learning rate of 0.01. This process enables the model to generate high-quality 128-dimensional descriptors}. The optimization of this stage relies on the two loss terms:  triplet loss $L_{T}$ and binarization loss $L_{B}$, which formulate the basic loss function as follows:
\begin{equation}
L_{basic}=L_{T} +\alpha_{B} L_{B},
\end{equation}
where $\alpha_{B}$ are weighting coefficients that control the binarization loss $L_{B}$.

In the second stage, we train the student model under the teacher's supervision, {\color{black}the learning rate is set to 0.01}. Therefore, the objective function includes both basic loss $L_{basic}$ and distillation loss $L_{distillation}$, it can be represented as:
\begin{equation} 
L_{train} = L_{basic} + \beta L_{distillation},
\end{equation}
where $\beta$ is the weighting coefficient of distillation loss {\color{black}, which is set to 2 during the training}.

\section{Experiment}
\begin{table*}[t!]
\begin{center}
\caption{The quantitative comparisons of different descriptors on UBC dataset. The values are FPR95.}
\label{FPR95}
\begin{tabular}{ccccccccc}
\hline
\multicolumn{1}{c|}{Train}                   & \multicolumn{1}{c|}{\multirow{2}{*}{\begin{tabular}[c]{@{}c@{}}Feature\\ Size\end{tabular}}} & Notredame      & Yosemite       & Liberty       & Yosemite      & Liberty        & \multicolumn{1}{c|}{Notredame}      & \multirow{2}{*}{Mean} \\ \cline{3-8}
\multicolumn{1}{c|}{Test}                    & \multicolumn{1}{c|}{}                                                                        & \multicolumn{2}{c}{Liberty}     & \multicolumn{2}{c}{Notredame} & \multicolumn{2}{c|}{Yosemite}                        &                       \\ \hline
\multicolumn{9}{c}{Traditional Methods}                                                                                                                                                                                                                                                      \\ \hline
\multicolumn{1}{c|}{SIFT}                    & \multicolumn{1}{c|}{4096-bit}                                                                & \multicolumn{2}{c}{29.84}       & \multicolumn{2}{c}{22.53}     & \multicolumn{2}{c|}{27.29}                           & 26.55                 \\
\multicolumn{1}{c|}{BRIEF}                   & \multicolumn{1}{c|}{256-bit}                                                                 & \multicolumn{2}{c}{59.15}       & \multicolumn{2}{c}{54.57}     & \multicolumn{2}{c|}{54.57}                           & 56.23                 \\
\multicolumn{1}{c|}{BinBoost}                & \multicolumn{1}{c|}{64-bit}                                                                  & 20.49          & 21.67          & 16.90         & 14.54         & 22.88          & \multicolumn{1}{c|}{18.97}          & 19.24                 \\
\multicolumn{1}{c|}{LDAHash}                 & \multicolumn{1}{c|}{128-bit}                                                                 & 49.66          & 49.66          & 51.58         & 51.58         & 52.95          & \multicolumn{1}{c|}{52.95}          & 51.40                 \\ \hline
\multicolumn{9}{c}{CNN Methods}                                                                                                                                                                                                                                                              \\ \hline
\multicolumn{1}{c|}{HardNet}                 & \multicolumn{1}{c|}{64-bit}                                                                  & 11.89          & 16.40          & 7.60          & 9.93          & 14.50          & \multicolumn{1}{c|}{13.20}          & 12.25                      \\
\multicolumn{1}{c|}{SOSNet}                  & \multicolumn{1}{c|}{64-bit}                                                                  & 11.26          & 15.95          & 7.04          & 9.24          & 13.64          & \multicolumn{1}{c|}{11.72}          & 11.48                 \\
\multicolumn{1}{c|}{HyNet}                   & \multicolumn{1}{c|}{64-bit}                                                                  & 11.72          & 16.16          & 9.94          & 9.03          & 13.14          & \multicolumn{1}{c|}{11.87}          & 11.97                 \\
\multicolumn{1}{c|}{Ours}                    & \multicolumn{1}{c|}{64-bit}                                                                  & 11.84             &   14.68    &10.14          &8.39          &       14.24         & \multicolumn{1}{c|}{11.58}               &11.81                       \\
\multicolumn{1}{c|}{Ours with distillation}  & \multicolumn{1}{c|}{64-bit}                                                                  & \textbf{10.74} & \textbf{13.40} & \textbf{6.66} & \textbf{8.26} & \textbf{11.95} & \multicolumn{1}{c|}{\textbf{11.29}} & \textbf{10.38}        \\ \hline
\multicolumn{9}{c}{CNN Methods with Data Augmentation}                                                                                                                                                                                                                                       \\ \hline
\multicolumn{1}{c|}{HardNet}                & \multicolumn{1}{c|}{64-bit}                                                                  & 11.38          & 16.03          & 6.89          & 8.79          & 14.13          & \multicolumn{1}{c|}{14.78}          & 12.00                 \\
\multicolumn{1}{c|}{SOSNet}                 & \multicolumn{1}{c|}{64-bit}                                                                  & 11.15          & 14.73          & 6.14          & 7.85          & 11.52          & \multicolumn{1}{c|}{11.85}          & 10.54                 \\
\multicolumn{1}{c|}{HyNet}                  & \multicolumn{1}{c|}{64-bit}                                                                  & 9.39           & 13.00          & 5.47          & 7.21          & 11.41          & \multicolumn{1}{c|}{10.56}          & 9.50                  \\
\multicolumn{1}{c|}{Ours}                   & \multicolumn{1}{c|}{64-bit}                                                                  & 9.62          & 12.52            & 8.43          & 7.12          & 11.56          & \multicolumn{1}{c|}{11.07}          &10.05               \\
\multicolumn{1}{c|}{Ours with distillation} & \multicolumn{1}{c|}{64-bit}                                                                  & \textbf{9.15}  & \textbf{10.87} & \textbf{5.24} & \textbf{6.50} & \textbf{10.57} & \multicolumn{1}{c|}{\textbf{10.23}} & \textbf{8.76}         \\ \hline
\end{tabular}
\end{center}
\vspace{-5mm}
\end{table*}

\begin{table}[t!]
\begin{center}
\caption{ The efficiency results of CNN methods.}
\label{runtime}
\begin{tabular}{|c|c|c|}
\hline
        & Parameters      & Runtime (ms)  \\ \hline
HardNet & 810272          & 20.50         \\ \hline
SOSNet  & 812067          & 25.64          \\ \hline
HyNet   & 812067          & 25.62          \\ \hline
Ours    & \textbf{578531} & \textbf{18.29} \\ \hline
\end{tabular}
\end{center}
\vspace{-5mm}
\end{table}

We perform two types of experiments to validate the proposed method.
First, we fairly compare the matching performance of our method and previous methods on the public UBC dataset~\cite{UBC}. Second, to demonstrate the application of our method to the MR-SLAM task, we build a MR-SLAM system based on VINS-Mono. With the system, we then conduct a MR-SLAM experiment on EuROC dataset~\cite{euroc}.   

\subsection{Patch matching}
\subsubsection{Dataset and implementation Details}
UBC dataset is the most widely used for patch matching~\cite{UBC}. It consists of three subsets: \emph{Liberty}, \emph{NotreDame}, and \emph{Yosemite}. The total dataset contains around 400k 64 $\times$ 64 patches with labels. Following the evaluation protocol, we train the models on one subset and test them on the other two subsets. Specifically, for every test subset, we pick 100k patches as the test set. Same as previous works, we show the matching performance by reporting the false positive rate at 95\% recall (FPR95). 

Several methods are taken as the baseline methods. The first type of methods are CNN-based methods, including HardNet~\cite{hardnet}, SOSNet~\cite{SOSnet}, and HyNet~\cite{HyNet}. 
{\color{black}To ensure fair comparisons, we set the output dimensions of our approach and all other methods to 64, by adjusting the number of output channels in their last layers
}. We report the matching results with and without data augmentation. Following the previous works, we augment the data by rotating part original patches before training.
The second type of methods are some traditional methods, including SIFT~\cite{SIFT}, BRIEF~\cite{BRIEF}, LDAHash~\cite{ldahash}, and BinBoost~\cite{binboost}. {\color{black}  We also report the parameter number and time consumption (of generating 500 descriptors) of each CNN method to visualize descriptor generation efficiency. All efficiency experiments are conducted on a desktop PC with an NVIDIA TITAN XP GPU. Furthermore, to test the generalization ability, we evaluate the performance of CNN methods trained on the HPatches dataset~\cite{hpatches} and tested on the UBC dataset.}
\subsubsection{Experimental results}

The comparison results of FPR95 are shown in Table \ref{FPR95}. 
It is clear that the CNN-based methods are significantly better than the traditional methods. On the other hand, matching results show that our method, even without distillation, is comparable to most CNN-based methods. Furthermore, with the assistance of the KD method, our network is able to outperform all baseline methods. The average FPR95 values of HardNet, SOSNet, HyNet, and our method (with KD) are 12.00, 10.54, 9.50, and 8.76, respectively. The lowest matching error is due to the consistency of our descriptors' distribution with high dimensional descriptors. The network efficiency results are shown in Table~\ref{runtime}. The results show that our model is around 30\% lighter than the baseline CNN models, and the running speed is also faster than the baseline methods. {\color{black} Table \ref{hpatches} further demonstrates our method's generalization ability, since  it achieves comparable performance to CNN methods despite being trained on a different dataset.}

{\color{black} }
\begin{table}[t!]
\begin{center}
\caption{Generalization performance (FPR95) of CNN methods ( trained on the HPatches dataset) on the UBC dataset.}
\label{hpatches}
\begin{tabular}{|c|c|c|}
\hline
           &UBC (Mean)  \\ \hline
HardNet    &14.36   \\ \hline
SOSNet          &13.94          \\ \hline
HyNet            &14.21           \\ \hline
Ours    & \textbf{13.21} \\ \hline
\end{tabular}
\end{center}
\vspace{-5mm}
\end{table}
\subsection{MR-SLAM Task}

\subsubsection{SLAM architecture}
\begin{figure}[t!]
\centering
\includegraphics[width=3in]{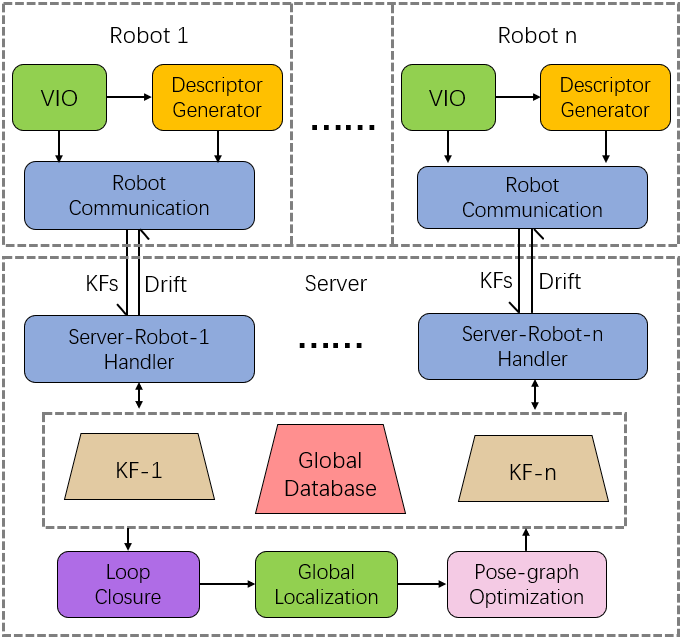}
\vspace{-2mm}
\centering
\caption{{\color{black}The framework of the MR-VINS system. All robots first perform VIO, which involves pure visual-based pose estimation and local bundle adjustment with IMU factors. Next, feature descriptors are generated and integrated into keyframes, which are transmitted to the server. Upon receiving the keyframe messages, the server executes BoW-based loop closure to detect loops in the robot trajectories. When loops are detected, the server performs global localization using PnP-RANSAC. Pose graph optimization is then carried out based on two factors: VIO and loop factors. Finally, the drift correction is sent back to the robots to correct real-time pose. }
}
\label{fig:SLAM}
\vspace{-3mm}
\end{figure}

To fairly test the descriptors' performance on the MR-SLAM task, we design a MR-SLAM system as the evaluation platform. To this end, we take VINS-Mono as our basic SLAM system and expand it into a multi-robots version, which we call the system MR-VINS. The detailed framework of MR-VINS is shown in Fig. \ref{fig:SLAM}. The whole system has a centralized architecture, i.e., the MR-SLAM system consists of several robots and a central server. We keep the visual-inertial odometry module on the robots while placing the back-end modules on the server. On the basis of this framework, we add the CNN descriptor models on the robots' side. Since the VINS-Mono exploits optical flow to track the features, the descriptors are not needed on the visual-inertial odometry module. Hence, in this experiment, we mainly apply the descriptors for loop closure and global localization. We place the descriptor models before the robot communication module so that the descriptors can be integrated into the keyframe message sent to the server.      
\subsubsection{Dataset and implementation Details}
For SLAM evaluation, we use the public EuRoC dataset~\cite{euroc}, which collects data from the Micro Aerial Vehicle (MAV). In our experiments, we evaluate the methods on the Machine Hall scene. {\color{black}Similar to the benchmark MR-SLAM's setting~\cite{CCM}, 
} we first conduct the experiment for SR-SLAM on sequences MH1, MH2, and MH3, respectively. Then, to test the performance under the MR-SLAM task, we take the experiment where three robots are run on these sequences (MH1-3) simultaneously.

The EuRoC dataset contains stereo images and IMU measurements. And in our experiments, we pick the left images and IMU measurements as the input of our SLAM system. 
We take two types of comparisons in the experiments. The first type is the comparison between different descriptor methods. We compare our method with previous CNN methods, including HardNet~\cite{hardnet}, SOSNet~\cite{SOSnet}, and HyNet~\cite{HyNet}. To have a fair comparison, we only change the descriptor extraction models in our MR-SLAM system. In addition, all the models are trained on HPatches dataset~\cite{hpatches}. The second type is the comparison between our method and benchmark MR-SLAM systems, including original MR-VINS (without CNN models) and {\color{black}CCM-SLAM~\cite{CCM} 
}. For evaluation, we take two measures: localization accuracy and communication bandwidth.
\subsubsection{Experimental results}

The quantitative results of the SLAM accuracy are shown in Table \ref{RMSE}. 
Compared to the original MR-VINS, as seen, CNN-based methods are able to improve the localization performance, especially for MR-SLAM. As the better descriptors detect more loops, particularly the ``hard'' inter-robot loops. {\color{black}Moreover, our method generally achieves comparable accuracy to the CNN methods on both single and multi-robots tasks. 
}However, since the SLAM performance depends on more than image matching. The advantages of our localization accuracy are not obvious as patch matching results.

While comparing our method against CCM-SLAM, we can see, CCM-SLAM achieves better performance on SR-SLAM tasks since it uses the bundle adjustment (BA) optimization method, which is able to produce more accurate results than the pose-graph method we use. On the other hand, our method has a more accurate result for MR-SLAM due to the much more detected inter-robot loops. Furthermore, low keyframe sending frequency also limits the accuracy of CCM-SLAM.

Table \ref{Bandwidth} shows the bandwidth of different SLAM systems. Due to the same dimension of the CNN-based descriptors, the bandwidth of them should be similar. Hence, we do not report the bandwidth of other CNN methods. As shown by the results, our method has a much lower bandwidth than the original MR-VINS and CCM-SLAM. As our message frequency is 3 times more than CCM-SLAM, the size of our keyframe massage is around 10 times smaller than CCM-SLAM. Narrow bandwidth is attributed to the extremely compact size of our descriptors (1/4 of normal BRIEF descriptors). In addition, due to the compact keyframe size, we do not need to reduce the keyframe frequency during the MR-SLAM task specifically. 
\begin{table}[t!]
\begin{center}
\caption{The Trajectory error (RMSE in cm) of Different methods.}
\label{RMSE}
\begin{tabular}{c|cccc|c}
\hline
                & \multicolumn{4}{c|}{SR-SLAM}                                  & MR-SLAM       \\ \cline{2-6} 
                & MH1          & MH2          & MH3          & Mean          & MH1-3       \\ \hline
CCM-SLAM        & \textbf{6.10} & 8.10          & \textbf{4.80} & \textbf{6.33} & 7.70          \\
MR-VINS         & 6.15          & 5.88          & 11.58         & 7.87          & 9.91          \\
MR-VINS+HardNet & 7.20          & 5.91          & 9.15          & 7.42          & 7.18          \\
MR-VINS+SOSNet  & 6.59          & 4.94          & 9.06          & 6.83          & 7.19          \\
MR-VINS+HyNet   & 6.48          & 5.44          & 8.43          & 6.78          & 7.08          \\
MR-VINS+Ours    & 6.26          & \textbf{4.14} & 9.41          & 6.60          & \textbf{6.94} \\ \hline
\end{tabular}
\vspace{-4mm}
\end{center}
\end{table}

\begin{table}[t!]
\begin{center}
\caption{The Bandwidth of Different methods.}
\label{Bandwidth}
\begin{tabular}{|c|c|c|}
\hline
Methods      & Frequency (Hz) & Bandwidth (kb/s) \\ \hline
CCM-SLAM     & 2              & 350              \\ \hline
MR-VINS      & 7              & 224              \\ \hline
MR-VINS+Ours & 7              & \textbf{119}     \\ \hline
\end{tabular}
\vspace{-5mm}
\end{center}
\end{table}

\section{CONCLUSIONS}
In this paper, we explore the problem of how to learn compact descriptors for MR-SLAM. We argue that there are two challenges. The first is how to achieve good matching when the descriptors are extremely compact. The second is the descriptor needs to be generated in real-time. To tackle these challenges, we propose a teacher-student framework that leverages a compact student model to estimate low-dimensional descriptors. As the output dimensions between the teacher and the student are different, we propose a distance-based distillation loss function that enables the knowledge distillation between different dimensional descriptors. Based on that, we are able to produce extremely compact descriptors with high performance. We fairly test our model on the public UBC dataset. As a result, our method outperforms baseline methods in both accuracy and efficiency. Moreover, we have developed a MR-SLAM system based on the proposed descriptor generation method. The results on the EuRoC dataset show that our method performs well in both accuracy and bandwidth on MR-SLAM tasks.
\bibliographystyle{IEEEtran}   
\bibliography{references.bib}

\begin{thebibliography}{10}
\providecommand{\url}[1]{#1}
\csname url@samestyle\endcsname
\providecommand{\newblock}{\relax}
\providecommand{\bibinfo}[2]{#2}
\providecommand{\BIBentrySTDinterwordspacing}{\spaceskip=0pt\relax}
\providecommand{\BIBentryALTinterwordstretchfactor}{4}
\providecommand{\BIBentryALTinterwordspacing}{\spaceskip=\fontdimen2\font plus
\BIBentryALTinterwordstretchfactor\fontdimen3\font minus
  \fontdimen4\font\relax}
\providecommand{\BIBforeignlanguage}[2]{{%
\expandafter\ifx\csname l@#1\endcsname\relax
\typeout{** WARNING: IEEEtran.bst: No hyphenation pattern has been}%
\typeout{** loaded for the language `#1'. Using the pattern for}%
\typeout{** the default language instead.}%
\else
\language=\csname l@#1\endcsname
\fi
#2}}
\providecommand{\BIBdecl}{\relax}
\BIBdecl

\bibitem{CCM}
P.~Schmuck and M.~Chli, ``Multi-uav collaborative monocular slam,'' in
  \emph{Proc. IEEE Int. Conf. Robot. Automat.}, 2017, pp. 3863--3870.

\bibitem{covins}
P.~Schmuck, T.~Ziegler, M.~Karrer, J.~Perraudin, and M.~Chli, ``Covins:
  Visual-inertial slam for centralized collaboration,'' in \emph{2021 IEEE
  International Symposium on Mixed and Augmented Reality Adjunct
  (ISMAR-Adjunct)}.\hskip 1em plus 0.5em minus 0.4em\relax IEEE, 2021, pp.
  171--176.

\bibitem{DoorSLAM}
P.-Y. Lajoie, B.~Ramtoula, Y.~Chang, L.~Carlone, and G.~Beltrame, ``Door-slam:
  Distributed, online, and outlier resilient slam for robotic teams,''
  \emph{IEEE Robotics and Automation Letters}, vol.~5, no.~2, pp. 1656--1663,
  2020.

\bibitem{XiyueGuo2021}
X.~Guo, J.~Hu, J.~Chen, F.~Deng, and T.~L. Lam, ``Semantic histogram based
  graph matching for real-time multi-robot global localization in large scale
  environment,'' \emph{IEEE Robotics and Automation Letters}, vol.~6, no.~4,
  pp. 8349--8356, 2021.

\bibitem{narrow1}
D.~Tardioli, E.~Montijano, and A.~R. Mosteo, ``Visual data association in
  narrow-bandwidth networks,'' in \emph{2015 IEEE/RSJ International Conference
  on Intelligent Robots and Systems (IROS)}.\hskip 1em plus 0.5em minus
  0.4em\relax IEEE, 2015, pp. 2572--2577.

\bibitem{narrow2}
D.~Van~Opdenbosch and E.~Steinbach, ``Collaborative visual slam using
  compressed feature exchange,'' \emph{IEEE Robotics and Automation Letters},
  vol.~4, no.~1, pp. 57--64, 2018.

\bibitem{L2-net}
Y.~Tian, B.~Fan, and F.~Wu, ``{L2-Net}: Deep learning of discriminative patch
  descriptor in euclidean space,'' in \emph{Proceedings of the IEEE Conference
  on Computer Vision and Pattern Recognition}, 2017, pp. 661--669.

\bibitem{hardnet}
A.~Mishchuk, D.~Mishkin, F.~Radenovic, and J.~Matas, ``Working hard to know
  your neighbor's margins: Local descriptor learning loss,'' \emph{Advances in
  Neural Information Processing Systems}, vol.~30, 2017.

\bibitem{SOSnet}
Y.~Tian, X.~Yu, B.~Fan, F.~Wu, H.~Heijnen, and V.~Balntas, ``Sosnet: Second
  order similarity regularization for local descriptor learning,'' in
  \emph{Proceedings of the IEEE/CVF Conference on Computer Vision and Pattern
  Recognition}, 2019, pp. 11\,016--11\,025.

\bibitem{HyNet}
Y.~Tian, A.~Barroso~Laguna, T.~Ng, V.~Balntas, and K.~Mikolajczyk, ``Hynet:
  Learning local descriptor with hybrid similarity measure and triplet loss,''
  \emph{Advances in Neural Information Processing Systems}, vol.~33, pp.
  7401--7412, 2020.

\bibitem{Hinton2015DistillingTK}
G.~E. Hinton, O.~Vinyals, and J.~Dean, ``Distilling the knowledge in a neural
  network,'' \emph{arXiv preprint arXiv:1503.02531}, 2015.

\bibitem{hu2021boosting}
J.~Hu, C.~Fan, H.~Jiang, X.~Guo, Y.~Gao, X.~Lu, and T.~L. Lam, ``Boosting
  light-weight depth estimation via knowledge distillation,'' \emph{arXiv
  preprint arXiv:2105.06143}, 2021.

\bibitem{wang2021knowledge}
L.~Wang and K.-J. Yoon, ``Knowledge distillation and student-teacher learning
  for visual intelligence: A review and new outlooks,'' \emph{IEEE Transactions
  on Pattern Analysis and Machine Intelligence}, vol.~44, no.~6, pp.
  3048--3068, 2021.

\bibitem{euroc}
M.~Burri, J.~Nikolic, P.~Gohl, T.~Schneider, J.~Rehder, S.~Omari, M.~W.
  Achtelik, and R.~Siegwart, ``The euroc micro aerial vehicle datasets,''
  \emph{The International Journal of Robotics Research}, vol.~35, no.~10, pp.
  1157--1163, 2016.

\bibitem{talk}
M.~Giamou, K.~Khosoussi, and J.~P. How, ``Talk resource-efficiently to me:
  Optimal communication planning for distributed loop closure detection,'' in
  \emph{2018 IEEE International Conference on Robotics and Automation
  (ICRA)}.\hskip 1em plus 0.5em minus 0.4em\relax IEEE, 2018, pp. 3841--3848.

\bibitem{near}
Y.~Tian, K.~Khosoussi, M.~Giamou, J.~P. How, and J.~Kelly, ``Near-optimal
  budgeted data exchange for distributed loop closure detection,'' \emph{arXiv
  preprint arXiv:1806.00188}, 2018.

\bibitem{object1}
J.~McCormac, R.~Clark, M.~Bloesch, A.~Davison, and S.~Leutenegger, ``Fusion++:
  Volumetric object-level slam,'' in \emph{2018 international conference on 3D
  vision (3DV)}.\hskip 1em plus 0.5em minus 0.4em\relax IEEE, 2018, pp. 32--41.

\bibitem{object2}
A.~Gawel, C.~Del~Don, R.~Siegwart, J.~Nieto, and C.~Cadena, ``X-view:
  Graph-based semantic multi-view localization,'' \emph{IEEE Robotics and
  Automation Letters}, vol.~3, no.~3, pp. 1687--1694, 2018.

\bibitem{SIFT}
D.~G. Lowe, ``Distinctive image features from scale-invariant keypoints,''
  \emph{International journal of computer vision}, vol.~60, no.~2, pp. 91--110,
  2004.

\bibitem{BRIEF}
M.~Calonder, V.~Lepetit, M.~Ozuysal, T.~Trzcinski, C.~Strecha, and P.~Fua,
  ``Brief: Computing a local binary descriptor very fast,'' \emph{IEEE
  Transactions on Pattern Analysis and Machine Intelligence}, vol.~34, no.~7,
  pp. 1281--1298, 2011.

\bibitem{PCASIFT}
Y.~Ke and R.~Sukthankar, ``Pca-sift: A more distinctive representation for
  local image descriptors,'' in \emph{Proceedings of the 2004 IEEE Computer
  Society Conference on Computer Vision and Pattern Recognition, 2004. CVPR
  2004.}, vol.~2.\hskip 1em plus 0.5em minus 0.4em\relax IEEE, 2004, pp.
  II--II.

\bibitem{ldahash}
C.~Strecha, A.~Bronstein, M.~Bronstein, and P.~Fua, ``Ldahash: Improved
  matching with smaller descriptors,'' \emph{IEEE transactions on Pattern
  Analysis and Machine Intelligence}, vol.~34, no.~1, pp. 66--78, 2011.

\bibitem{binboost}
T.~Trzcinski, M.~Christoudias, P.~Fua, and V.~Lepetit, ``Boosting binary
  keypoint descriptors,'' in \emph{Proceedings of the IEEE Conference on
  Computer Vision and Pattern Recognition}, 2013, pp. 2874--2881.

\bibitem{matchnet}
X.~Han, T.~Leung, Y.~Jia, R.~Sukthankar, and A.~C. Berg, ``Matchnet: Unifying
  feature and metric learning for patch-based matching,'' in \emph{Proceedings
  of the IEEE Conference on Computer Vision and Pattern Recognition}, 2015, pp.
  3279--3286.

\bibitem{unsuper1}
K.~Lin, J.~Lu, C.-S. Chen, and J.~Zhou, ``Learning compact binary descriptors
  with unsupervised deep neural networks,'' in \emph{Proceedings of the IEEE
  Conference on Computer Vision and Pattern Recognition}, 2016, pp. 1183--1192.

\bibitem{unsuper2}
Y.~Duan, Z.~Wang, J.~Lu, X.~Lin, and J.~Zhou, ``Graphbit: Bitwise interaction
  mining via deep reinforcement learning,'' in \emph{Proceedings of the IEEE
  Conference on Computer Vision and Pattern Recognition}, 2018, pp. 8270--8279.

\bibitem{unsuper3}
M.~Zieba, P.~Semberecki, T.~El-Gaaly, and T.~Trzcinski, ``Bingan: Learning
  compact binary descriptors with a regularized gan,'' \emph{Advances in Neural
  Information Processing Systems}, vol.~31, 2018.

\bibitem{hu2022data}
J.~Hu, C.~Fan, M.~Ozay, H.~Jiang, and T.~L. Lam, ``Data-free dense depth
  distillation,'' \emph{arXiv preprint arXiv:2208.12464}, 2022.

\bibitem{hu2022progressive}
J.~Hu, C.~Fan, M.~Ozay, H.~Feng, Y.~Gao, and T.~L. Lam, ``Progressive
  self-distillation for ground-to-aerial perception knowledge transfer,''
  \emph{arXiv preprint arXiv:2208.13404}, 2022.

\bibitem{Mobilenets}
A.~G. Howard, M.~Zhu, B.~Chen, D.~Kalenichenko, W.~Wang, T.~Weyand,
  M.~Andreetto, and H.~Adam, ``Mobilenets: Efficient convolutional neural
  networks for mobile vision applications,'' \emph{arXiv preprint
  arXiv:1704.04861}, 2017.

\bibitem{UBC}
M.~Brown, G.~Hua, and S.~Winder, ``Discriminative learning of local image
  descriptors,'' \emph{IEEE Transactions on Pattern Analysis and Machine
  Intelligence}, vol.~33, no.~1, pp. 43--57, 2010.

\bibitem{hpatches}
V.~Balntas, K.~Lenc, A.~Vedaldi, and K.~Mikolajczyk, ``Hpatches: A benchmark
  and evaluation of handcrafted and learned local descriptors,'' in
  \emph{Proceedings of the IEEE Conference on Computer Vision and Pattern
  Recognition}, 2017, pp. 5173--5182.

\end{thebibliography}

\end{document}